\documentclass[journal]{IEEEtran}
\usepackage{lineno,hyperref}
\modulolinenumbers[5]
\usepackage{bm}
\usepackage{comment}
\usepackage{changepage}
\usepackage{amsmath}
\usepackage{graphicx}
\usepackage{subfigure}
\usepackage{multirow}
\usepackage{array}
\newcolumntype{P}[1]{>{\centering\arraybackslash}p{#1}}
\usepackage{makecell}
\usepackage{pgfplots}
\usepackage{dingbat}
\usepackage[flushleft]{threeparttable}
\usepackage{multirow}
\usepackage{cite}
\definecolor{darkgreen}{rgb}{0.0, 0.5, 0.0}
\begin{document}
\title{Semantic Segmentation of Remote Sensing Images with Sparse Annotations}


\author{Yuansheng~Hua,
        Diego~Marcos,
        Lichao~Mou, %~\IEEEmembership{Member,~IEEE,}
        Xiao Xiang Zhu,~\IEEEmembership{Fellow, ~IEEE },
        Devis~Tuia,~\IEEEmembership{Senior Member,~IEEE}
        % <-this % stops a space
\thanks{YH, LM and XXZ are with Data Science in Earth Observation, Technical University of Munich, Germany, and Remote Sensing Technology Institute, German Aerospace Center, Germany. (e-mails: yuansheng.hua@dlr.de; lichao.mou@dlr.de; xiaoxiang.zhu@dlr.de) DM is/DT was with the Laboratory of GeoInformation Science and Remote Sensing, Wageningen University, the Netherlands. (e-mail: diego.marcos@wur.nl). He is now with the Ecole Polytechnique F{\'e}d{\'e}rale de Lausanne, Sion, Switzerland. (e-mail: devis.tuia@epfl.ch)% <-this % stops a space
\textit{(Correspondences: Xiao Xiang Zhu, Devis Tuia.)}}
\thanks{The work is supported by the German Federal Ministry of Education and Research -- AI future lab "AI4EO" (Grant number: 01DD20001).}}

\newpage
\thispagestyle{empty}
\onecolumn
\noindent This work has been submitted to the IEEE for possible publication. Copyright may be transferred without notice, after which this version may no longer be accessible.
\newpage
\twocolumn

\markboth{IEEE Geoscience and Remote Sensing Letters}%
{Hua \MakeLowercase{\textit{et al.}}: Learning with Sparse Scribbled Annotations}
\maketitle

\begin{abstract}
\textcolor{blue}{This is the preprint version. To read the final
version, please go to IEEE Geoscience and Remote Sensing Letters.} Training Convolutional Neural Networks (CNNs) for very high resolution images requires a large quantity of high-quality pixel-level annotations, which is extremely labor- and time-consuming to produce. Moreover, professional photo interpreters might have to be involved for guaranteeing the correctness of annotations. To alleviate such a burden, we propose a framework for semantic segmentation of aerial images based on incomplete annotations, where annotators are asked to label a few pixels with easy-to-draw scribbles. To exploit these sparse scribbled annotations, we propose the FEature and Spatial relaTional regulArization (FESTA) method to complement the supervised task with an unsupervised learning signal that accounts for neighbourhood structures both in spatial and feature terms. For the evaluation of our framework, we perform experiments on two remote sensing image segmentation datasets involving aerial and satellite imagery, respectively. Experimental results demonstrate that the exploitation of sparse annotations can significantly reduce labeling costs while the proposed method can help improve the performance on semantic segmentation when training on such annotations. The sparse labels and codes are publicly available for reproducibility purposes\footnote{https://github.com/Hua-YS/Semantic-Segmentation-with-Sparse-Labels}.
\end{abstract}

\begin{IEEEkeywords}
Semantic segmentation, aerial image, sparse scribbled annotation, convolutional neural networks, semi-supervised learning.
\end{IEEEkeywords}

\IEEEpeerreviewmaketitle
\vspace{-0.5cm}
\section{Introduction}
\begin{figure}[t]
    \centering
    \subfigure[]{\includegraphics[width=0.075\textwidth]{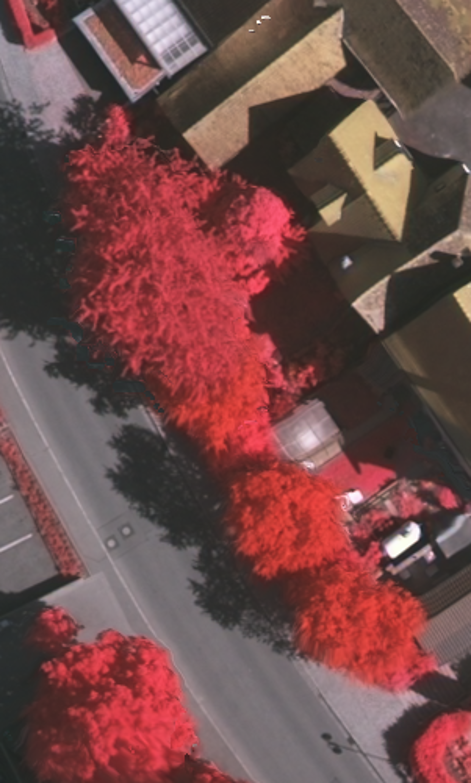}\label{fig:comp_image}}
    \hspace{-0.2em}
    \subfigure[]{\includegraphics[width=0.075\textwidth]{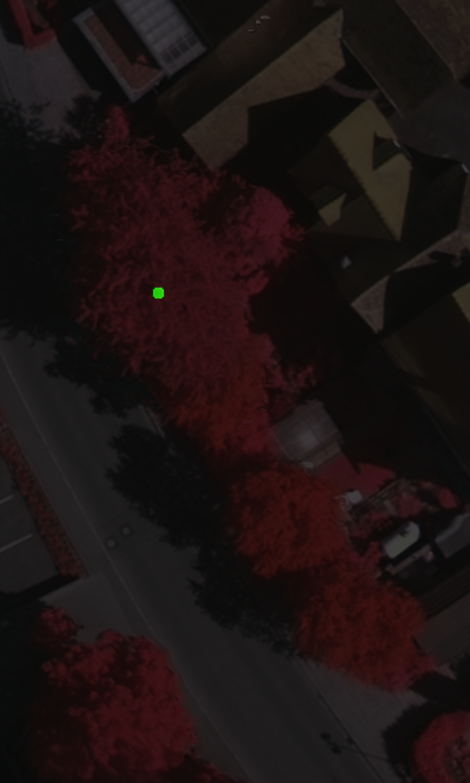}\label{fig:comp_point}}
    \hspace{-0.2em}
    \subfigure[]{\includegraphics[width=0.075\textwidth]{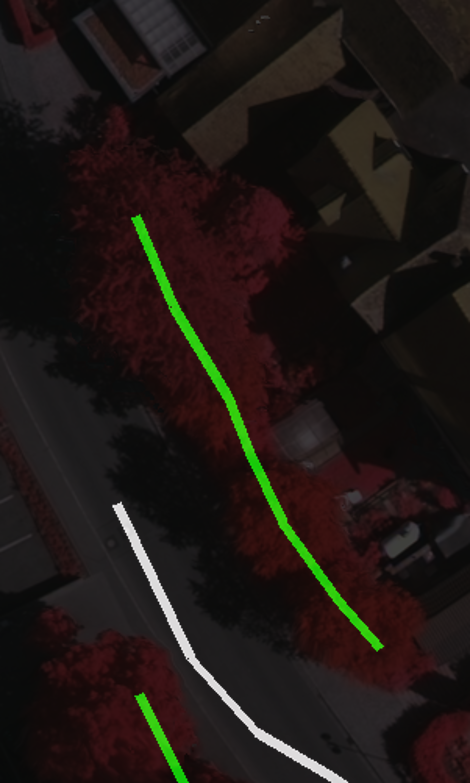}\label{fig:comp_line}}
    \hspace{-0.2em}
    \subfigure[]{\includegraphics[width=0.075\textwidth]{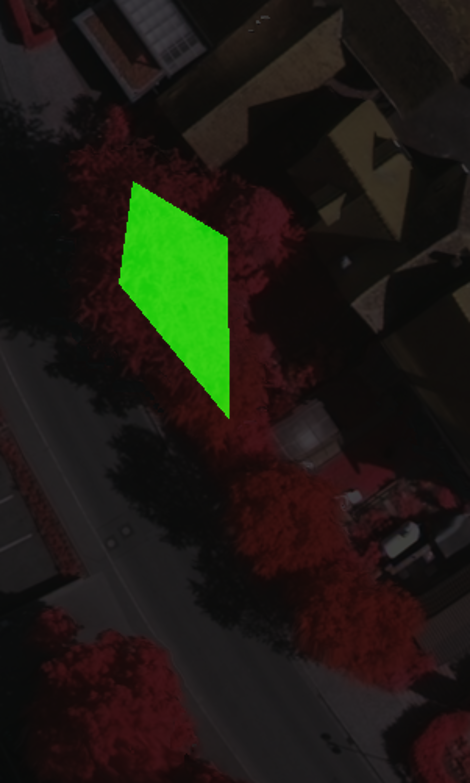}\label{fig:comp_polygon}}
    \hspace{-0.2em}
    \subfigure[]{\includegraphics[width=0.075\textwidth]{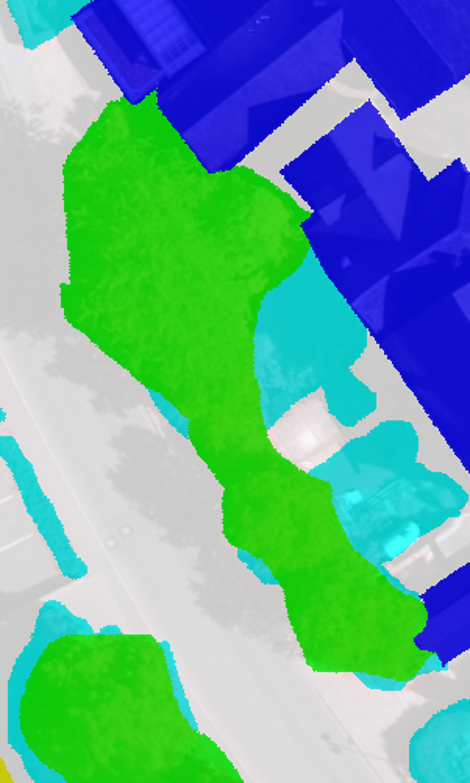}}
    \caption{Comparisons of different levels of scribbled annotations. Trees (marked as \textcolor{green}{green}) are taken as an example here. Images from left to right are (a) an aerial image, (b) point-, (c) line- and (d) polygon-level scribbled annotations, and (e) dense pixel-wise labels.}
    \label{fig:scribble_comparison}
    \vspace{-0.8cm}
\end{figure}

Semantic segmentation of remote sensing imagery aims at identifying the land-cover or land-use category of each pixel in an image. As one of the fundamental visual tasks, semantic segmentation has been attracting wide attention in the remote sensing community and proven to be beneficial to a variety of applications, such as land cover mapping, traffic monitoring and urban management. Recently, many studies~\cite{zhu2017deep} resort to learning deep Convolutional Neural Networks (CNNs) with full supervision for semantic segmentation and have obtained enormous achievements. However, training a fully supervised segmentation CNN requires a huge volume of dense pixel-level ground truths, which are labor- and time-consuming to generate. Moreover, expert annotators might be needed for correctly identifying pixels located at object boundaries and ambiguous regions (e.g., shadows in Fig.~\ref{fig:scribble_comparison}).

To alleviate the requirement of dense pixel-wise annotations, semi-supervised learning approaches are proposed to make use of additional information, such as spatial relations (e.g. neighboring pixels are likely to belong to the same class) or feature-level relations (e.g. pixels with similar CNN feature representations are likely to belong to the same class), for semantic segmentation. These methods aim to utilize low-cost annotations, such as points~\cite{bearman2016s}, scribbles~\cite{wu2018scribble,maggiolo2018improving} or image-level labels~\cite{nivaggioli2019weakly,zhu2019semi}. As the first attempt, Bearman~\emph{et al}.~\cite{bearman2016s} proposed to learn semantic segmentation models with point-level supervision, where only one point is labeled for each instance. In scribble-supervised algorithms, annotations are provided in the form of hand-drawn scribbles. Wu~\emph{et al}.~\cite{wu2018scribble} propose to learn aerial building footprint segmentation models from scribbles. Maggiolo~\emph{et al}. \cite{maggiolo2018improving} argue that a network directly trained on scribbled ground truths fails to accurately predict object boundaries and propose to employ a fully connected Conditional Random Field (CRF) to refine the shapes of objects. Compared to fully annotated ground truths, scribbled annotations (cf., Fig.~\ref{fig:comp_line}) are easier to generate in a user-friendly way. In comparison with point-level annotations (e.g., Fig.~\ref{fig:comp_point}), scribbles can provide stronger supervisory signals. However, point- and scribble-supervised segmentation methods remain under-explored in the remote sensing community. To this end, we propose a simple yet effective framework for semantic segmentation of remote sensing imagery with low-cost annotations. In this framework, we manually create point- or scribble-level annotations and train networks on them. Besides, we also evaluate polygon-level annotations (see Fig.~\ref{fig:comp_polygon}), which can be easily yielded and cover more pixels than the other types of annotations. Since these annotations are sparsely distributed across the images, we call them sparse annotations in the following sections. In order to better exploit sparse annotations, we propose a semi-supervised learning method which encodes and regularizes the feature and spatial relations. To demonstrate the effectiveness of our learning framework, extensive experiments are conducted on two VHR datasets, the Vaihingen and Zurich Summer.

\section{Methodology}
\subsection{Supervision with Sparse Annotations}

In contrast to conventional dense annotations, sparse annotations have two characteristics: 1) a very small proportion pixels are assigned semantic classes, and 2) objects do not need to be entirely annotated (see (b), (c), (d) in Fig.~\ref{fig:scribble_comparison}). This greatly reduces the effort required from the annotators, as complex boundaries and ambiguous pixels can be avoided.

Here we consider three levels of sparse annotations: point-, scribble-, and polygon-level. Specifically, point-level annotations indicate that, for an annotator interaction, only one single pixel is labeled. Scribble-level annotations, also called line-level annotations, are yielded by drawing a scribble line within an object and assigning all pixels along this line the same class label. Similarly, polygon-level annotations can be generated by drawing a polygon within an object and classifying pixels located in the polygon into the same semantic class. Examples of these three levels of annotations are shown in Fig.~\ref{fig:scribble_comparison}.

%To summarize, scribbled annotations enjoy two advantages: 1) With a few clicks or casual drawing, scribbled annotations can be easily produced, which might cost only several seconds for labeling one object. 2) Annotators are not required to be armed with comprehensive knowledge of candidate classes, as they merely need to label pixels they are confident with. We note that in the field of semantic aerial image segmentation, all these scribbled annotations are largely unexplored. [Diego: this paragraph sounds like discussion, not methodology]

\subsection{Feature and Spatial Relational Regularization}

\begin{figure}[t]
    \centering
    \includegraphics[width=0.3\textwidth]{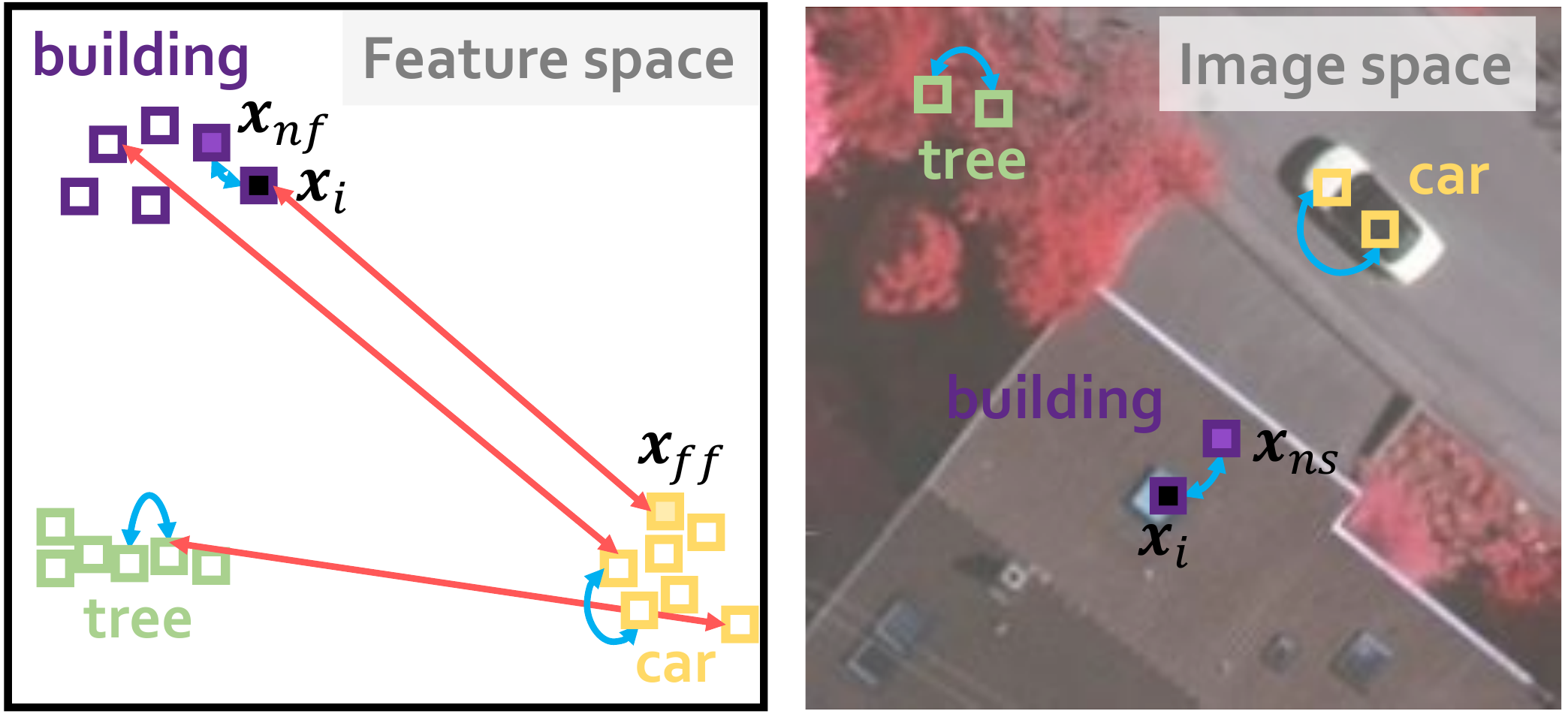}
    \caption{Illustration of the proposed FESTA. A Sample $\bm{x}_i$ belonging to \textit{building} (filled with black) is taken as an example.}
    \label{fig:FESTA}
    \vspace{-0.5cm}
\end{figure}

When using sparse annotations, the vast majority of pixels in the training images are left unlabelled.
In order to exploit both labeled and unlabeled pixels, we develop a semi-supervised methodology, named FEature and Spatial relaTional regulArization (FESTA), to enable a semantic segmentation CNN to learn discriminative features, while leveraging the unlabelled image pixels. An assumption shared by many unsupervised learning algorithms~\cite{cover1967nearest} is that nearby entities often belong to the same class. Based on this assumption, a recent work~\cite{sabokrou2019self} achieves success in representation learning by encoding neighborhood-relations in the feature space. Inspired by this work, we propose to encode and regularize relations between pixels in both feature and spatial domain, as shown in Fig.~\ref{fig:FESTA}, so that the learned features become more useful for semantic segmentation.

Specifically, given a sample $\bm{x}_i$ (\emph{i.e.}, a CNN feature vector extracted from location $i$ in an image), we first encode its relations to all other samples by measuring the distance in space and feature similarity with respect to all other features in the image.
The sample with the smallest similarity is considered as the far-away sample in the feature space, $\bm{x}_{i_{ff}}$, while that with the highest similarity is defined as the neighboring sample in feature space, $\bm{x}_{i_{nf}}$. 
According to the aforementioned proximity assumption, it is highly probable that $\bm{x}_i$ and $\bm{x}_{i_{nf}}$ belong to the same class, and thus, the distance between them should be as small as possible. In order to prevent a trivial solution in which all features collapse to the same point, $\bm{x}_i$ and $\bm{x}_{i_{ff}}$ are encouraged to further increase their dissimilarity. We apply a similar reasoning in the spatial domain, since images are smooth in spatial terms. Thus, we take the 8 spatial neighbors of $\bm{x}_i$ into consideration and chose the one most similar in feature space as the spatial neighbor, $\bm{x}_{i_{ns}}$. This operation is intended to prevent pairing $\bm{x}_i$ with a spatial neighbor that belongs to the object boundary. %Besides, encoding and regularizing spatial relations is beneficial to identifying classes with complicated textures as illustrated in Fig.~\ref{fig:FESTA}. Without any doubt, $\bm{x}_{i_{ns}}$ is supposed to be as close as possible to $\bm{x}_i$. [Diego: I don't get this sentece, so I'd rather remove it]

These priors can be incorporated into the learning objectives by using the following loss function:
\begin{equation}
\begin{aligned}
    \mathcal{L}_{FESTA} = &\hspace{0.4em}\alpha\sum\limits_{i=1}^N\mathcal{D}(\bm{x}_i, \bm{x}_{i_{nf}})+\beta\sum\limits_{i=1}^N\mathcal{D}(\bm{x}_i, \bm{x}_{i_{ns}})\\
     &\hspace{0.2em}+\gamma\sum\limits_{i=1}^N\mathcal{S}(\bm{x}_i, \bm{x}_{i_{ff}}),
    %\mathcal{L}_{FESTA} = \alpha\sum\limits_{i=1}^N\left\Vert\bm{x}_i-    \bm{x}_{i_{nf}}\right\Vert+\beta\sum\limits_{i=1}^N\left\Vert\bm{x}_i- \bm{x}_{i_{ns}}\right\Vert+\gamma\sum\limits_{i=1}^N\frac{\bm{x}_i \cdot\bm{x}_{i_{ff}}}{||\bm{x}_i||||\bm{x}_{i_{ff}}||},
\end{aligned}
\end{equation}
where $\mathcal{D}$ denotes the euclidean distance and $\mathcal{S}$ represents cosine similarity. $\alpha$, $\beta$, and $\gamma$ are trade-off parameters representing the significances of the respective terms, and $N$ represents the number of pixels in a given image. By minimizing $\mathcal{L}_{FESTA}$, $\bm{x}_{i_{nf}}$ and $\bm{x}_{i_{ns}}$ are forced to move closer to $\bm{x}_i$, while $\bm{x}_{i_{ff}}$ is pushed far from $\bm{x}_i$. In order to jointly exploit the sparse scribbled annotations and FESTA for the network training, the final loss is defined as:
\begin{equation}
\label{eq:final}
    \mathcal{L} = \mathcal{L}_{ce} +\lambda\mathcal{L}_{FESTA},
\end{equation}
where $\mathcal{L}_{ce}$ indicates the categorical cross-entropy loss calculated from pixels with annotations.

\subsection{CRF for Boundary Refinement}
To further refine the predictions of networks trained on scribbled annotations, we integrate a fully connected CRF~\cite{krahenbuhl2011efficient} into our system, and the energy function of CRF model is
\begin{equation}
    E = \sum\limits_{i}\theta_u(x_i) + \sum\limits_{ij}\theta_p(x_i, x_j),
\end{equation}
where $\theta_u(x_i)$ is the unary potential and calculated as $\theta_u(x_i)=-\log P(x_i)$. Here $i$ ranges from $0$ to the number of pixels in the image, and $P(x_i)$ is the label probability of pixel $i$. $\theta_p(x_i, x_j)$ is utilized to measure pairwise potentials between pixel $i$ and $j$. We tested with two Gaussian kernels,
\begin{equation}
\label{eq:dcrf}
    \begin{array}{l}
    k_1 = {\rm exp}(-\frac{||p_i-p_j||^2}{2\theta_1^2}-\frac{||I_i-I_j||^2}{2\theta_2^2}), \\
    k_2 = {\rm exp}(-\frac{||p_i-p_j||^2}{2\theta_3^2}),
    \end{array}
\end{equation}
where $p_i$ and $I_i$ indicate the position and color intensity of pixel $i$. $\theta_1$, $\theta_2$, and $\theta_3$ are hyperparameters that control the kernel "scale". In Eq.~\ref{eq:dcrf}, $k_1$ is known as \textit{appearance kernel} and tends to classify neighboring pixels with similar appearances~\cite{schindler2012}, i.e., color intensities, into the same classes, while $k_2$, so-called \textit{smoothness kernel}, penalizes pixels nearby but assigned diverse labels. This step is expected to make the class map smoother within homogeneous areas.

\section{Experimental Results}
\subsection{Dataset Description}

The Vaihingen dataset\footnote{\url{http://www2.isprs.org/commissions/comm3/wg4/2d-sem-label-vaihingen.html}} is a benchmark dataset for semantic segmentation provided by the International Society for Photogrammetry and Remote Sensing (ISPRS). 33 aerial images with a spatial resolution of 9 cm were collected over the city of Vaihingen, and each image covers an average area of 1.38 km$^2$. %Sizes of images are not uniform, and an average size is $2494 \times 2064$ pixels. 
For each aerial image, three bands are available, near infrared (NIR), red (R), and green (G). Besides, coregistered digital surface models (DSMs) are provided for all images. 16 images are fully annotated. In total, six land-cover classes are considered: impervious surface, building, low vegetation, tree, car, and clutter/background. In this paper, we follow the train-test split scheme in most existing works~\cite{Maggiori2017HighResolutionAI, sherrah2016fully} and select five images (image IDs: 11, 15, 28, 30, 34) as the test set. The remaining ones are utilized to train our models. 

The Zurich Summer dataset~\cite{volpi2015semantic} is composed of 20 images, which are taken over the city of Zurich in August 2002 by the QuickBird satellite. The spatial resolution is 0.62 m, and the average size of images is $1,000 \times 1,150$ pixels. The images consist of four channels: near infrared (NIR), red (R), green (G), and blue (B). Following previous works~\cite{tuia2018decision, wendl2019novelty}, we only utilize NIR, R, and G in our experiments and train our model on 15 images; the others (image IDs: 16, 17, 18, 19, 20) are utilized to test. In total, there are 8 urban classes, including road, building, tree, grass, bare soil, water, railway, and swimming pool. Uncategorized pixels are labeled as background. 

It is noteworthy that although full pixel-wise annotations are provided for all images in the Vaihingen and Zurich Summer dataset, we only use them in the test phase to calculate evaluation metrics. The training of all models is done with scribbled annotations described below.

\begin{table}[!t]
\renewcommand{\arraystretch}{1.1}
\centering
\caption{The total numbers of pixels labeled with sparse point-, line-, and polygon-level annotations (middle three columns) and dense annotations (right column) in the Vaihingen and Zurich Summer datasets.}
\label{tab:total_dataset}
%\begin{tabular}{P{4cm}|P{1.2cm}P{1.2cm}|P{1.8cm}|P{1.8cm}}
\begin{threeparttable}
\begin{tabular}{c|ccc|c}
\Xhline{2\arrayrulewidth}
Dataset Name & Point & Line & Polygon & Dense* \\% & S/F\\
\hline
\hline
Vaihingen & 18,787 & 480,593 & 4,591,409 & 54,373,518 \\
\hline
Zurich Summer & 29,508 & 330,767 & 1,445,270 & 12,266,287 \\
\Xhline{2\arrayrulewidth}
\end{tabular}
\begin{tablenotes}
\item *Background/Clutter is not considered. 
\end{tablenotes}
\end{threeparttable}
\vspace{-0.5cm}
\end{table}

\subsection{Scribbled Annotation Generation}
\label{sec:annotate_scribble}
\begin{figure}[t]
    \centering
    \includegraphics[width=.35\textwidth]{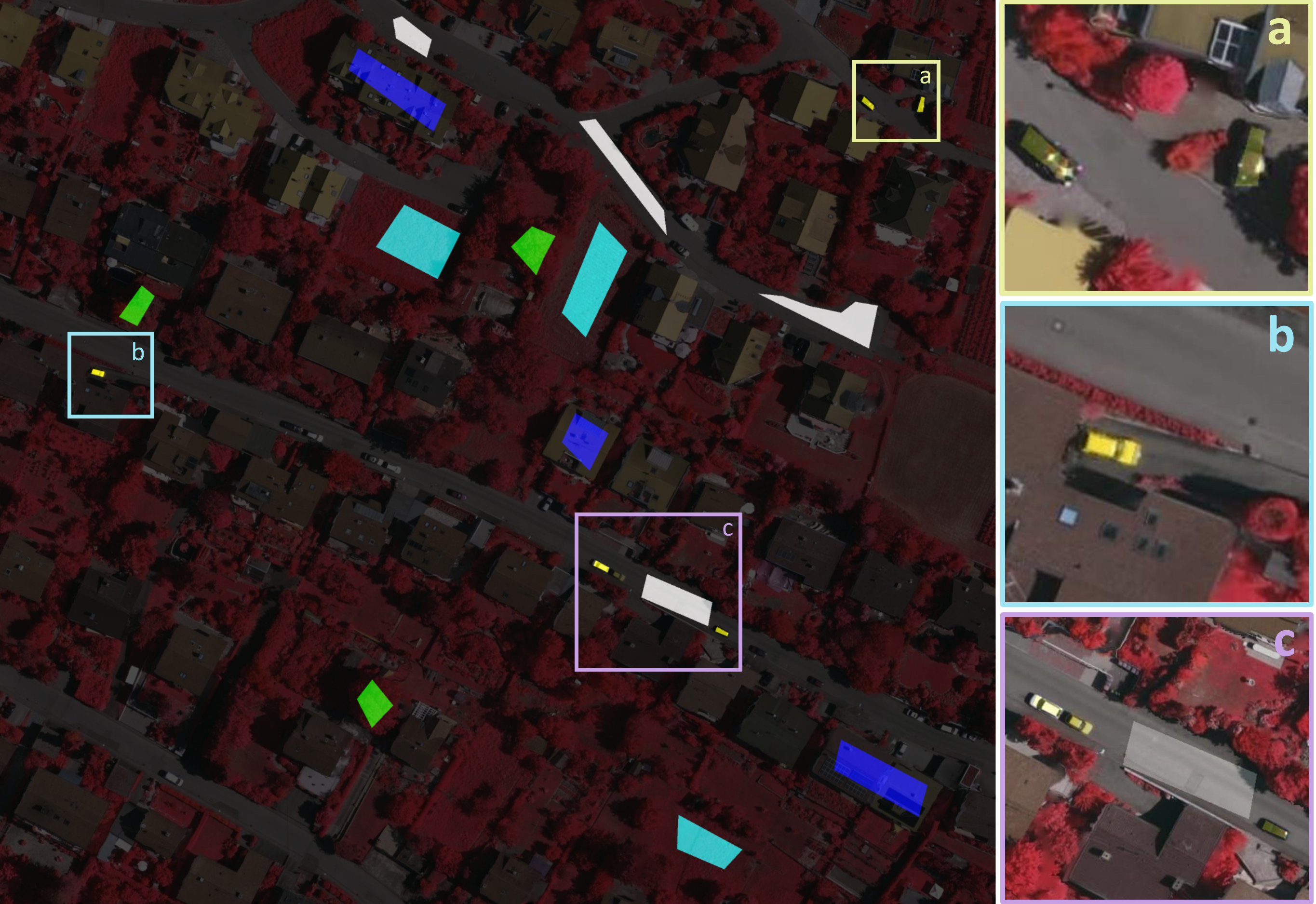}
    \vspace{-0.3cm}
    \caption{Example polygon-level annotations of an image (ID: 13) on the Vaihingen dataset. Annotations of cars are zoomed in to illustrate that annotations should include variant visual appearances for one class. Legend---\textcolor{pink}{white}: impervious surfaces, \textcolor{blue}{blue}: buildings, \textcolor{cyan}{cyan}: low vegetation, \textcolor{green}{green}: trees, \textcolor{yellow}{yellow}: cars.}
    \label{fig:scribble}
    \vspace{-0.5cm}
\end{figure}

To annotate large-scale images, we employ an online labeling platform, LabelMe~\footnote{\url{http://labelme.csail.mit.edu/Release3.0/}}, and ask annotators to draw by following these rules: 1) for each class, annotations are supposed to cover diverse appearances (see region a, b, and c in Figure~\ref{fig:scribble}, where cars of different colors are annotated) and be located in different positions of the image separately. 2) polygon- and line-level annotations are not required to delineate object boundaries precisely, see the annotations of trees in Fig.~\ref{fig:comp_line} and \ref{fig:comp_polygon}. In order to make the time spent on each level of scribbled annotations more equivalent, we ask 4 annotators (including 2 non-experts) to label 7, 5, and 3 objects per class for point-, line- and polygon-level annotations in each aerial image. As a consequence, sparse but accurate annotations can be provided rapidly without effort. Since a point- or line-level annotation is often located in the centre area of an object and distant from its boundary, we perform morphological dilation on all point- and line-level annotations with a disk of radius 3. Afterwards, pixels involved in dilated annotations are assigned the same class labels as their central points or lines. For polygon-level annotations, pixels within each polygon are assigned the corresponding classes.

\begin{table*}[!t]
\scriptsize
\centering
\caption{Numerical results on the Vaihingen dataset (\%): We show the per-class $F_1$ score, mean $F_1$ score, and overall accuracy on the test set. Mean and standard deviation of each metric are calculated from results on sparse annotations produced by 4 annotators. Results on dense annotations are provided as reference.}
\vspace{-0.4em}
\label{tab:vaihingen_result}
\centering
\begin{tabular}{c|c|ccccc|c|c}
\Xhline{2\arrayrulewidth}
Scribble & Model & Imp. surf. & Build. & Low veg. & Tree & Car & mean $F_1$ & OA \\
\hline
\hline
\multirow{4}{*}{Point} & FCN-WL~\cite{cciccek20163d} & 69.81 $\pm$ 1.52 & 75.02 $\pm$ 2.32 & 60.25 $\pm$ 3.40  & 76.17 $\pm$ 1.42 & 12.29 $\pm$ 3.60 & 58.71 $\pm$ 0.33 & 67.11 $\pm$ 0.97\\
& FCN+dCRF~\cite{maggiolo2018improving} & 75.37 $\pm$ 0.93 & \textbf{81.37} $\pm$ 3.10 & \textbf{61.93} $\pm$ 5.54 & \textbf{78.50} $\pm$ 1.69 & 17.51 $\pm$ 6.70 & 62.94 $\pm$ 0.44 & 72.53 $\pm$ 0.42\\
& FCN-FESTA & 74.65 $\pm$ 2.73 & 78.64 $\pm$ 4.74 & 60.24 $\pm$ 3.33 & 76.15 $\pm$ 2.07 & 23.65 $\pm$ 4.24 & 62.66 $\pm$ 2.54 & 71.43 $\pm$ 2.93 \\
& FCN-FESTA+dCRF & \textbf{77.62} $\pm$ 1.93 & 80.08 $\pm$ 5.27 & 60.78 $\pm$ 4.00 & 76.70 $\pm$ 2.00 & \textbf{31.40} $\pm$ 5.24 & \textbf{65.32} $\pm$ 2.56 & \textbf{73.65} $\pm$ 2.52 \\
\hline
\multirow{4}{*}{Line} & FCN-WL~\cite{cciccek20163d} & 78.44 $\pm$ 3.24 & 83.45 $\pm$ 1.58 & 64.02 $\pm$ 2.34 & 79.32 $\pm$ 0.54 & 29.01 $\pm$ 2.96 & 66.85 $\pm$ 1.81 & 76.12 $\pm$ 1.52 \\
& FCN+dCRF~\cite{maggiolo2018improving} & \textbf{81.32} $\pm$ 2.45 & \textbf{84.88} $\pm$ 1.88 & 63.71 $\pm$ 3.92 & 79.88 $\pm$ 1.33 & 38.95 $\pm$ 4.50 & 69.75 $\pm$ 2.23 & \textbf{78.03} $\pm$ 1.82 \\
& FCN-FESTA & 78.12 $\pm$ 3.92 & 83.76 $\pm$ 2.00 & \textbf{65.78} $\pm$ 1.88 & \textbf{80.49} $\pm$ 0.93 & 38.24 $\pm$ 10.31 & 69.28 $\pm$ 3.66 & 77.24 $\pm$ 2.27 \\
& FCN-FESTA+dCRF & 80.06 $\pm$ 3.32 & 84.47 $\pm$ 2.23 & 64.35 $\pm$ 2.38 & 80.32 $\pm$ 0.92 & \textbf{43.72} $\pm$ 9.62 & \textbf{70.58} $\pm$ 3.42 & 77.99 $\pm$ 2.14 \\
\hline
\multirow{4}{*}{Polygon} & FCN-WL~\cite{cciccek20163d} & 76.71 $\pm$ 3.63 & 80.03 $\pm$ 1.42 & 59.40 $\pm$ 6.09 & 78.50 $\pm$ 2.86 & 26.28 $\pm$ 11.06 & 64.19 $\pm$ 4.40 & 74.18 $\pm$ 2.97 \\
& FCN+dCRF~\cite{maggiolo2018improving} & 78.37 $\pm$ 3.08 & 80.85 $\pm$ 1.13 & 57.92 $\pm$ 7.67 & 78.67 $\pm$ 2.87 & 29.13 $\pm$ 8.15 & 64.99 $\pm$ 3.99 & 75.15 $\pm$ 2.94 \\
& FCN-FESTA & 78.98 $\pm$ 3.82 & 83.10 $\pm$ 2.62 & \textbf{62.59} $\pm$ 4.89 & \textbf{79.91} $\pm$ 3.31 & 33.04 $\pm$ 7.71 & 67.52 $\pm$ 4.07 & 76.65 $\pm$ 3.39 \\
& FCN-FESTA+dCRF & \textbf{80.62} $\pm$ 3.22 & \textbf{83.62} $\pm$ 2.29 & 60.79 $\pm$ 5.04 & 79.81 $\pm$ 2.52 & \textbf{40.27} $\pm$ 8.30 & \textbf{69.02} $\pm$ 4.01 & \textbf{77.32} $\pm$ 2.92 \\
\hline
Dense & FCN~\cite{long2015fully} & 88.67  & 92.83  & 76.32  & 74.21  & 86.67  & 83.74  & 86.51  \\
\Xhline{2\arrayrulewidth}
\end{tabular}
\end{table*}

\begin{table*}[!t]
\vspace{-0.3cm}
\scriptsize
\centering
\caption{Numerical results on the Zurich Summer dataset (\%): We show the per-class $F_1$ score, mean $F_1$ score, and overall accuracy on the test set. Mean and standard deviation of each metric are calculated from results on sparse annotations produced by 4 annotators. Results on dense annotations are provided as reference.}
\vspace{-0.4em}
\label{tab:zurich_result}
\centering
\begin{tabular}{P{0.75cm}|P{2.03cm}|P{1.02cm}P{1.02cm}P{1.02cm}P{1.02cm}P{1.02cm}P{1.02cm}P{0.9cm}P{1.15cm}|P{1.06cm}|P{1.0cm}}
\Xhline{2\arrayrulewidth}
Scribble & Model & Road & Build. & Tree & Grass & Soil & Water & Rail. & Pool & mean $F_1$ & OA  \\
\hline
\hline
\multirow{4}{*}{Point} & FCN-WL~\cite{cciccek20163d} & 69.74$\pm$3.98 & 78.94$\pm$3.01 & 82.33$\pm$2.55 & 82.20$\pm$2.40 & 53.37$\pm$7.03 & 87.87$\pm$1.40 & 0.81$\pm$1.42 & 48.89$\pm$9.42 & 63.02$\pm$2.14 & 77.38$\pm$2.73 \\
& FCN+dCRF~\cite{maggiolo2018improving} & \textbf{72.13}$\pm$4.99 & \textbf{80.71}$\pm$1.84 & 82.87$\pm$2.08 & 83.55$\pm$2.07 & 63.92$\pm$8.90 & 92.71$\pm$1.26 & \textbf{2.09}$\pm$4.17 & 59.96$\pm$14.60 & 67.24$\pm$1.93 & \textbf{80.03}$\pm$2.26 \\
& FCN-FESTA & 70.64$\pm$3.44 & 77.34$\pm$4.13 & \textbf{82.91}$\pm$2.48 & 83.73$\pm$2.34 & 56.67$\pm$5.64 & 89.67$\pm$2.25 & 0.94$\pm$1.89 & 73.62$\pm$4.06 & 66.94$\pm$2.56 & 78.17$\pm$3.00 \\
 & FCN-FESTA+dCRF & 71.23$\pm$2.61 & 77.71$\pm$3.17 & 82.81$\pm$1.99 & \textbf{84.18}$\pm$1.96 & \textbf{66.34}$\pm$3.69 & \textbf{93.40}$\pm$1.81 & 0.00$\pm$0.00 & \textbf{77.38}$\pm$8.87 & \textbf{69.05}$\pm$1.15 & 79.11$\pm$2.14 \\
\hline
\multirow{3}{*}{Line} & FCN-WL\cite{cciccek20163d} & 73.00$\pm$4.60 & \textbf{81.17}$\pm$3.77 & \textbf{82.82}$\pm$2.78 & 81.88$\pm$1.41 & 67.02$\pm$8.77 & 90.98$\pm$1.79 & 1.19$\pm$1.60 & 58.77$\pm$7.82 & 67.10$\pm$2.02 & \textbf{79.75}$\pm$2.25 \\
& FCN+dCRF~\cite{maggiolo2018improving} & 71.71$\pm$4.83 & 79.22$\pm$4.01 & 81.22$\pm$3.06 & 80.43$\pm$2.10 & \textbf{71.72}$\pm$9.20 & 84.65$\pm$14.90 & \textbf{2.35}$\pm$4.71 & 67.58$\pm$17.39 & 68.39$\pm$3.10 & 78.84$\pm$2.15 \\
& FCN-FESTA & \textbf{73.34}$\pm$3.88 & 79.08$\pm$3.60 & 82.71$\pm$2.10 & \textbf{84.27}$\pm$1.41 & 60.67$\pm$13.36 & 92.37$\pm$1.44 & 1.02$\pm$0.83 & 74.27$\pm$8.24 & 68.47$\pm$2.45 & 79.52$\pm$2.86 \\
& FCN-FESTA+dCRF & 71.74$\pm$2.78 & 75.81$\pm$4.18 & 81.20$\pm$1.60 & 83.44$\pm$1.51 & 66.49$\pm$15.57 & \textbf{94.68}$\pm$0.52 & 0.00$\pm$0.00 & \textbf{82.06}$\pm$6.80 & \textbf{69.43}$\pm$2.57 & 78.51$\pm$2.21 \\
\hline
\multirow{3}{*}{Polygon} & FCN-WL~\cite{cciccek20163d} & 64.18$\pm$6.14 & 72.17$\pm$6.01 & 79.64$\pm$4.25 & 77.10$\pm$3.92 & 49.17$\pm$16.96 & 89.26$\pm$3.52 & 1.31$\pm$1.09 & 76.90$\pm$6.33 & 63.72$\pm$4.35 & 73.09$\pm$4.49 \\
& FCN+dCRF~\cite{maggiolo2018improving} & 62.63$\pm$5.77 & 70.35$\pm$4.88 & 78.30$\pm$3.53 & 75.94$\pm$4.42 & 52.11$\pm$14.06 & 91.03$\pm$4.39 & 0.84$\pm$1.69 & \textbf{85.13}$\pm$2.72 & 64.54$\pm$4.08 & 72.37$\pm$3.89 \\
& FCN-FESTA & \textbf{66.53}$\pm$5.07 & \textbf{74.06}$\pm$3.06 & \textbf{80.05}$\pm$3.66 & \textbf{79.42}$\pm$3.56 & 57.83$\pm$11.38 & 90.80$\pm$2.42 & 5.87$\pm$4.86 & 65.68$\pm$16.06 & 65.03$\pm$1.98 & \textbf{75.00}$\pm$3.17 \\
& FCN-FESTA+dCRF & 65.10$\pm$4.42 & 71.96$\pm$2.76 & 79.44$\pm$3.26 & 78.87$\pm$4.58 & \textbf{61.86}$\pm$9.72 & \textbf{92.50}$\pm$2.96 & \textbf{6.37}$\pm$6.63 & 77.21$\pm$6.63 & \textbf{66.66}$\pm$2.41 & 74.41$\pm$2.86 \\
\hline
Dense & FCN~\cite{long2015fully} & 88.34 & 93.27 & 92.40 & 89.48 & 67.96 & 96.87 & 2.98 & 88.10 & 77.42 & 90.51 \\
\Xhline{2\arrayrulewidth}
\end{tabular}
\vspace{-0.4cm}
\end{table*}

Table~\ref{tab:total_dataset} shows the average amounts of pixels with sparse and dense annotations in both datasets. It can be seen that sparse annotations are several orders of magnitude fewer than dense annotations. As to the labeling time, it took on average $133$, $126$, and $161$ seconds per image to produce point-, line- and polygon-level annotations, respectively, for the Vaihingen dataset, and $177$, $162$, and $238$ seconds per image for the Zurich Summer dataset. In Section \ref{sec:experiments}, we demonstrate the proposed method allows to improve the semantic segmentation results using these sparse annotations. In Section \ref{sec:experiments_more}, we discuss the differences observed among the tested annotation types.

\subsection{Training Details}
We segment the images with a standard FCN (i.e., FCN-16s~\cite{long2015fully}) and initialize convolutional layers with Glorot uniform~\cite{glorot2010understanding} initializers. Specifically, VGG-16 is taken as the backbone, and outputs of the last two convolutional blocks are upsampled to the original resolution and fused with an element-wise addition. The fused feature maps are finally fed into a convolutional layer, where the number of filters is equivalent to the number of classes. In the training phase, all weights are trainable and updated with Nestrov Adam~\cite{nadam2}, using $\beta_1=0.9$, $\beta_2=0.999$, and $\epsilon=1\mathrm{e}{-08}$ as recommended. We initialize the learning rate as $2\mathrm{e}{-04}$ and let it decay by a factor of 10 when validation loss is saturated. To train the network, we define the loss as Eq.~\ref{eq:final}, and $\lambda$ is set experimentally to 0.1 and 0.01 for the Vaihingen and Zurich Summer datasets, respectively. Tradeoff parameters, $\alpha$, $\beta$, and $\gamma$, are set as 0.5, 1.5, and 1, to ensure that 1) the regularizers governing feature and spatial relations are balanced, and 2) neighboring pixels in the image space receive more attention. The network, as well as FESTA, is implemented on TensorFlow and trained on one NVIDIA Tesla P100 16GB GPU for 100k iterations. The size of mini-batch is set as 5 during the training procedure. In the training phase, we use a sliding window to crop training images into $256 \times 256$ patches, and its stride is set to 64 pixels. Besides, no class-dependent configurations are considered. In the test phase, we employ dense CRF to refine predictions before calculating metrics. We tuned the parameters of dense CRF ($\theta_1$, $\theta_2$, and $\theta_3$ in Eq. \ref{eq:dcrf}) on validation images, and find that satisfactory results can be achieved for both FCN and FCN-FESTA when setting them to 30, 10, and 10, respectively. In the case of large homogeneous areas of an image belonging to the same class, $\alpha$ should be set to a small value, which encourages the network to focus more on geographically nearby samples. Besides, a large batch size and sliding window can also help alleviate the influence of such a scenario.

\subsection{Comparing with Existing Methods}
\label{sec:experiments}
We compare a Fully Convolutional Network~\cite{long2015fully} (FCN) learned using the proposed FESTA (FCN-FESTA) against an FCN learned with weighted loss function (FCN-WL)~\cite{cciccek20163d} on sparse annotations. We also report segmentation results of the baseline FCN trained on dense labels.
In addition, we study the influence of the fully connected CRF by comparing FCN-FESTA+dCRF and FCN+dCRF~\cite{maggiolo2018improving}. Each model is trained and validated on sparse annotations independently. Per-class $F_1$ scores, mean $F_1$ scores, and overall accuracy (OA) are calculated on test images with dense annotations. Considering that each model is learned on labels from four annotators, respectively, we average metrics obtained by each annotator and report them in the form of mean $\pm$ standard deviation.

\begin{figure}[!t]
%\captionsetup[subfigure]{labelformat=empty}
\centering
\subfigure{\includegraphics[width=0.080\textwidth]{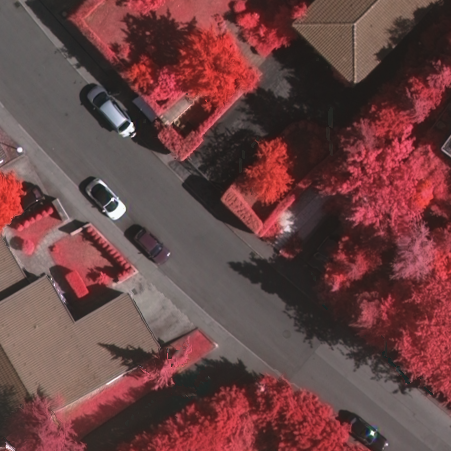}}
\hspace{-0.4em}
\subfigure{\includegraphics[width=0.080\textwidth]{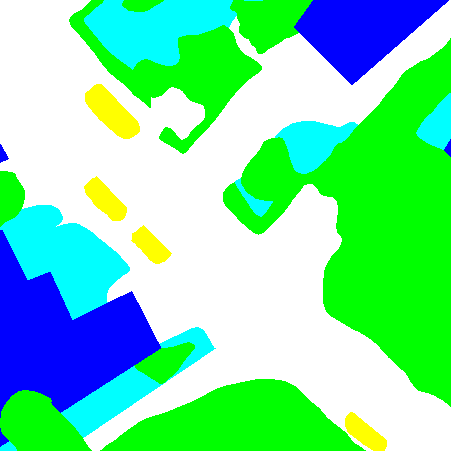}}
\hspace{-0.4em}
\subfigure{\includegraphics[width=0.080\textwidth]{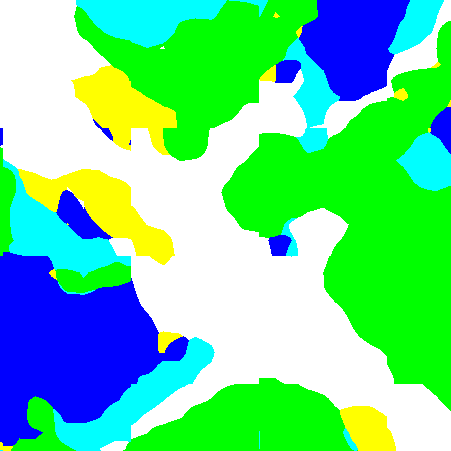}}
\hspace{-0.4em}
\subfigure{\includegraphics[width=0.080\textwidth]{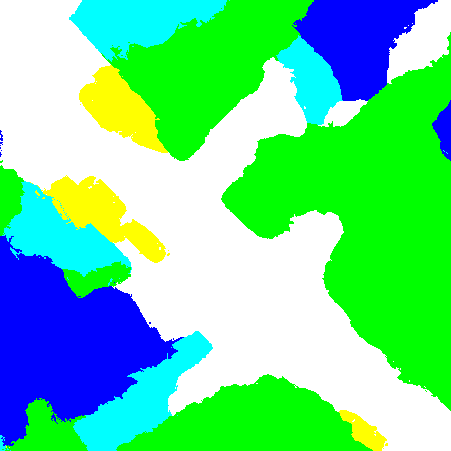}}
\hspace{-0.4em}
\subfigure{\includegraphics[width=0.080\textwidth]{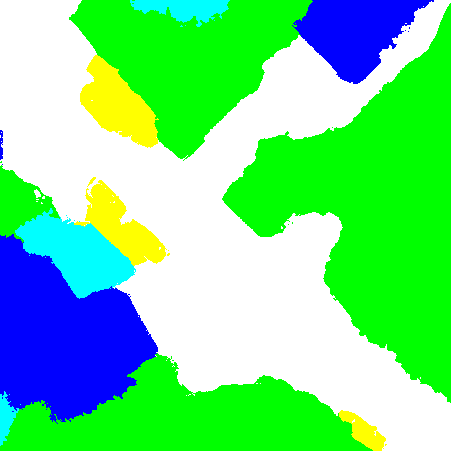}}
\vspace{-0.3em}
    
\subfigure{\includegraphics[width=0.080\textwidth]{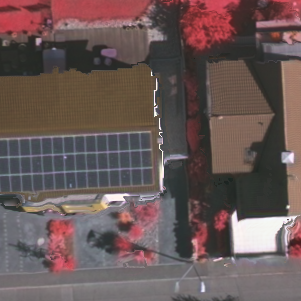}}
\hspace{-0.4em}
\subfigure{\includegraphics[width=0.080\textwidth]{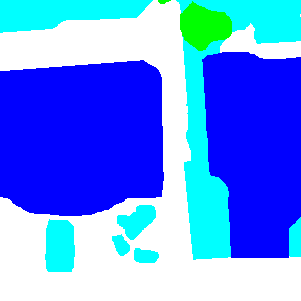}}
\hspace{-0.4em}
\subfigure{\includegraphics[width=0.080\textwidth]{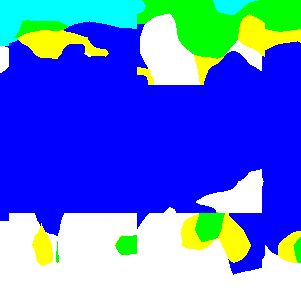}}
\hspace{-0.4em}
\subfigure{\includegraphics[width=0.080\textwidth]{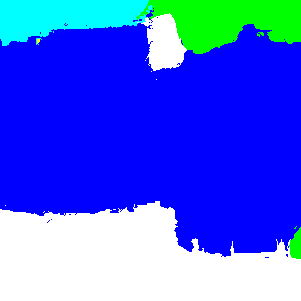}}
\hspace{-0.4em}
\subfigure{\includegraphics[width=0.080\textwidth]{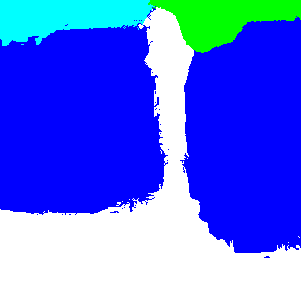}}
\vspace{-0.3em}
   
\addtocounter{subfigure}{-10} 
\subfigure[~image]{\includegraphics[width=0.080\textwidth]{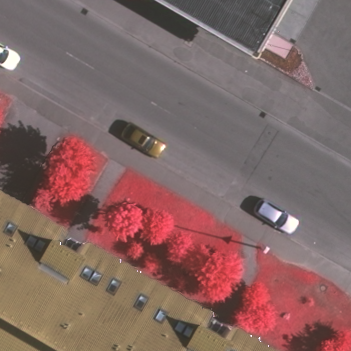}}
\hspace{-0.4em}
\subfigure[~{\scriptsize dense} {\tiny GT}]{\includegraphics[width=0.080\textwidth]{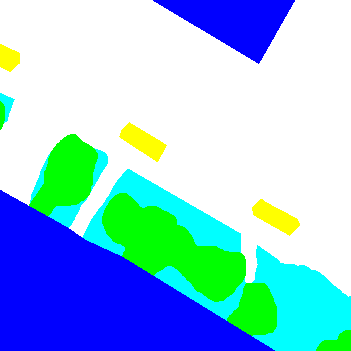}}
\hspace{-0.4em}
\subfigure[~{\tiny FCN-WL}]{\includegraphics[width=0.080\textwidth]{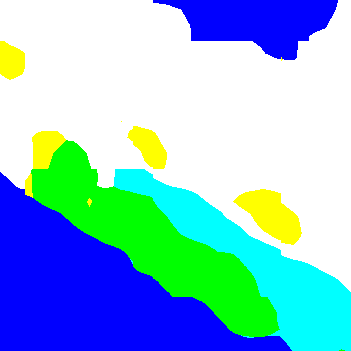}}
\hspace{-0.4em}
\subfigure[~{\tiny FCN}{\tiny +dCRF}]{\includegraphics[width=0.080\textwidth]{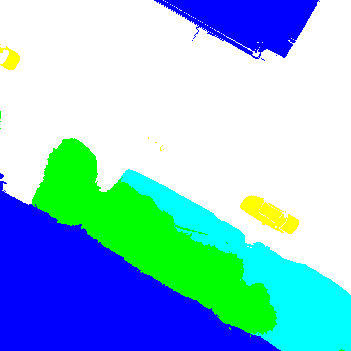}}
\hspace{-0.4em}
\subfigure[~ours]{\includegraphics[width=0.080\textwidth]{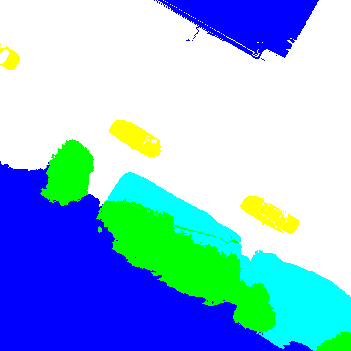}}
\vspace{-0.2em}

\caption{Examples of segmentation results on the Vaihingen dataset. All models are trained on line annotations. The legend is the same as that in Fig.~\ref{fig:scribble}.}
\label{fig:visual_vaihingen}
\end{figure}

Table~\ref{tab:vaihingen_result} exhibits numerical results on the Vaihingen dataset. FCN-FESTA+dCRF achieves the highest mean $F_1$ scores in training with all kinds of scribbled annotations, which demonstrates its effectiveness. To be more specific, with point- and polygon-level supervision, FCN-FESTA improves the mean $F_1$ score by 3.95\% and 3.33\% compared to FCN-WL, respectively. By refining predictions with dense CRF, FCN-FESTA+dCRF achieves improvements of 2.38\% and 4.03\% in comparison with FCN+dCRF. It is interesting to observe that line-level scribbles improve the segmentation performance the most, and FCN-FAST+dCRF learned with such annotations obtains the highest mean $F_1$ score, 70.58\%. Moreover, we note that FESTA can enhance the network capability of recognizing small objects, i.e., \textit{car}, in high resolution aerial images. Example segmentation results of networks trained on line annotations are visualized in Fig.~\ref{fig:visual_vaihingen}.

\begin{figure}[!t]
\vspace{-0.4cm}
%\captionsetup[subfigure]{labelformat=empty, captionskip=2pt}
\centering
\subfigure{\includegraphics[width=0.080\textwidth]{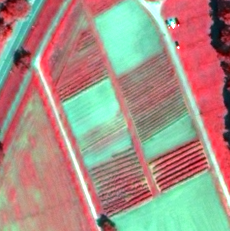}}
\hspace{-0.4em}
\subfigure{\includegraphics[width=0.080\textwidth]{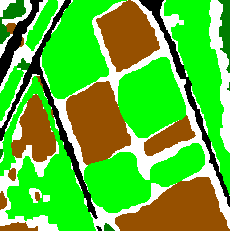}}
\hspace{-0.4em}
\subfigure{\includegraphics[width=0.080\textwidth]{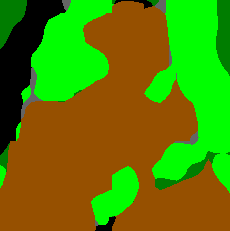}}
\hspace{-0.4em}
\subfigure{\includegraphics[width=0.080\textwidth]{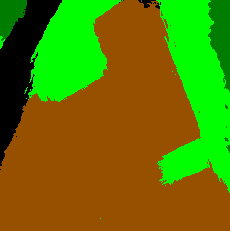}}
\hspace{-0.4em}
\subfigure{\includegraphics[width=0.080\textwidth]{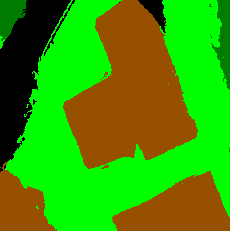}}
\vspace{-0.3em}

\subfigure{\includegraphics[width=0.080\textwidth]{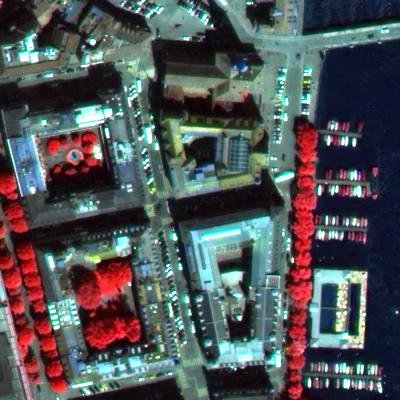}}
\hspace{-0.4em}
\subfigure{\includegraphics[width=0.080\textwidth]{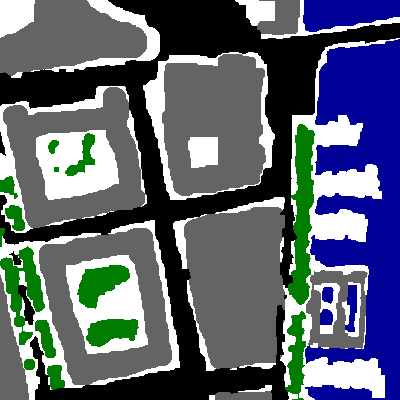}}
\hspace{-0.4em}
\subfigure{\includegraphics[width=0.080\textwidth]{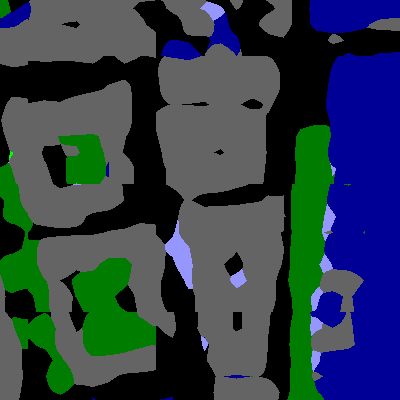}}
\hspace{-0.4em}
\subfigure{\includegraphics[width=0.080\textwidth]{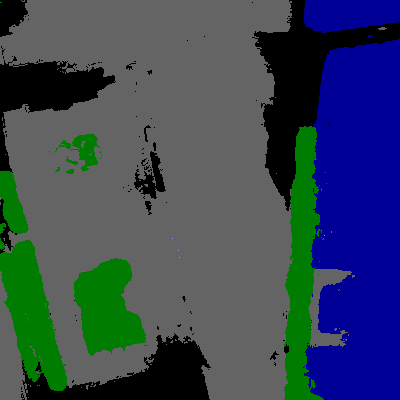}}
\hspace{-0.4em}
\subfigure{\includegraphics[width=0.080\textwidth]{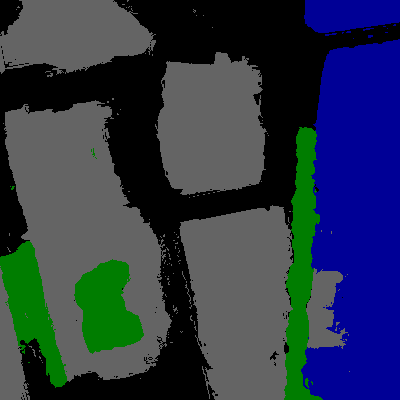}}
\vspace{-0.3em}

\addtocounter{subfigure}{-10} 
\subfigure[~image]{\includegraphics[width=0.080\textwidth]{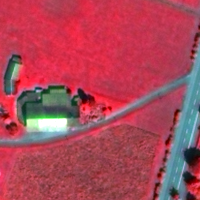}}
\hspace{-0.4em}
\subfigure[~{\scriptsize dense} {\tiny GT}]{\includegraphics[width=0.080\textwidth]{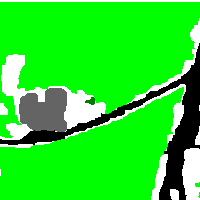}}
\hspace{-0.4em}
\subfigure[~{\tiny FCN-WL}]{\includegraphics[width=0.080\textwidth]{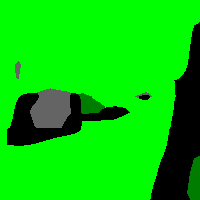}}
\hspace{-0.4em}
\subfigure[~{\tiny FCN}{\tiny +dCRF}]{\includegraphics[width=0.080\textwidth]{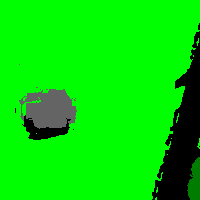}}
\hspace{-0.4em}
\subfigure[~ours]{\includegraphics[width=0.080\textwidth]{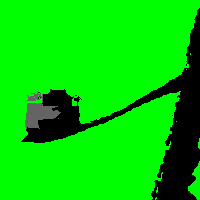}}
\vspace{-0.2em}

\caption{Examples of segmentation results on the Zurich Summer dataset. All models are trained on line annotations. Legend---black: road, \textcolor{brown}{brown}: soil, \textcolor{green}{green}: grass, \textcolor{darkgreen}{dark green}: tree, \textcolor{gray}{gray}: building, and \textcolor{pink}{white}: background.}
\label{fig:visual_zurich}
\vspace{-0.5cm}
\end{figure}

Numerical results on the Zurich Summer dataset are shown in Table~\ref{tab:zurich_result}. As can be seen, FESTA contributes to increments of 3.92\%, 1.37\% and 1.31\% in the mean $F_1$ score when training with point-, line- and polygon-level annotations. By utilizing line annotations and dense CRF, FCN-FESTA+dCRF obtains the highest mean $F_1$ score, 69.43\%. Besides, we note that the exploitation of dense CRF plays a significant role in improving results of networks trained on point-level scribbles. Example visual results of networks trained on line annotations are shown in Fig.~\ref{fig:visual_zurich}. In our experiments, we also train networks with multi-class dice loss and find that results are comparative to those learned with crossentropy loss.

\label{sec:experiments_more}
%\vspace{-4cm}

\subsection{Discussion on annotation type}

To further study the influence of annotations, we also train baseline FCNs on dense annotations and report numerical results in Tables~\ref{tab:vaihingen_result} and \ref{tab:zurich_result}. As shown in Tables~\ref{tab:vaihingen_result} and \ref{tab:zurich_result}, line-level annotations lead to the best performance on both datasets, even though the number of labeled pixels is an order of magnitude smaller than polygon annotations (see Table~\ref{tab:total_dataset}). Although it was expected that line annotations would outperform point annotations, due to their ability to capture within-object variations, we were surprised to see that they also outperformed polygon annotations. We suspect that this is linked to the fact that the number of pixels per object grows quadratically for polygons and linearly for lines. This would lead to a more balanced weighing of differently sized objects in the case of line annotations and an under-weighing of smaller objects in the case of polygon annotations, which could harm the model’s performance. Another reason could be that, since drawing a line is faster than drawing a polygon, annotators for the line features provided more scribbles in the same time budget.

In spite of the mean $F_1$ performance boost provided by FESTA, there is still a large gap with respect to the FCN model trained with dense ground truths, of 13\% in Vaihingen and 8\% in Zurich. This gap is, however, not evenly distributed across the classes. The gap is smaller or non-existent in classes such as water, tree, grass or soil, which are often homogeneous in terms of materials. On the contrary, it is larger for classes with more diverse materials (and therefore observed spectral values), such as building and car (in the Vaihingen dataset). It is noteworthy to mention that the class railway, in the Zurich dataset, is systematically missed in all cases, including the densely supervised FCN.

\section{Conclusion}
In this paper, we propose a simple yet efficient framework for semantic aerial image segmentation using sparse annotations and a semi-supervised learning objective. In order to validate the effectiveness of our approach, we conduct experiments on the Vaihingen and Zurich Summer datasets. Numerical and visual results suggest that the proposed method contributes to the improvement of semantic segmentation results using several kinds of sparse annotations. Although models learned on sparse annotations achieve relatively lower accuracies than those using dense annotations, we show that using a semi-supervised deep learning approach can help closing this performance gap while leveraging sparse annotations that can significantly reduce the costs of label generation. As future work, the proposed framework can be further improved by introducing graph-based models and prior knowledge learned from label semantics.

\section*{Acknowledgment}
The authors would like to thank Yingya Xu, Li Hua, and Yanping Tang for contributing to this work with annotation.

\ifCLASSOPTIONcaptionsoff
  \newpage
\fi

\bibliographystyle{IEEEtran}
\tiny
\bibliography{reference}

%\end{IEEEbiography}

\end{document}